\documentclass[a4paper, 12pt]{article}
\usepackage[margin=1in]{geometry}
\usepackage{times}

\usepackage{titling}
\pretitle{\vspace*{-4em}\begin{center} \fontsize{14}{16.8} \textbf{i-SAIRAS 2024}\\[6ex]}
\posttitle{\\[6ex]\end{center}}

\preauthor{%
  \normalsize
  \noindent
}
\postauthor{%
    \\[-8ex]
}

\usepackage{sectsty}
\sectionfont{\fontsize{12}{15}\selectfont}
\subsectionfont{\fontsize{12}{15}\selectfont}

\usepackage[
    backend=bibtex,
    style=numeric,
    maxbibnames=15,
    giveninits=true,
    doi=false,
    isbn=false,
    url=false,
]{biblatex}
\AtEveryBibitem{\clearfield{series}}
\AtEveryBibitem{\clearfield{urlyear}}

\usepackage{graphicx}
\usepackage{cleveref}

\crefformat{footnote}{#2\footnotemark[#1]#3}

\usepackage{etoolbox}
\patchcmd{\thebibliography}
  {\settowidth}
  {\setlength{\parsep}{0pt}\setlength{\itemsep}{0pt plus 0.1pt}\settowidth}
  {}{}

\makeatletter
\renewenvironment{abstract}{%
      \begin{center}%
        {\bfseries\normalsize\abstractname\vspace{\z@}}
      \end{center}%
      \quotation
    }
    \endquotation
\makeatother

\newcommand{\tss}[1]{\textsuperscript{#1}}
\newcommand{\abreak}{\\[1ex]}

\title{\textbf{The Importance of Adaptive Decision-Making for \\[0.5ex]Autonomous Long-Range Planetary Surface Mobility}}

\author{\tss{1*}Olivier Lamarre, \tss{2}Jonathan Kelly \abreak
\tss{*}lead presenter \abreak
\tss{1} olivier.lamarre@robotics.utias.utoronto.ca, Space and Terrestrial Autonomous Robotic \mbox{Systems} (STARS) Laboratory, University of Toronto Institute for Aerospace Studies, Canada \abreak
\tss{2} jonathan.kelly@robotics.utias.utoronto.ca, Space and Terrestrial Autonomous Robotic \mbox{Systems} (STARS) Laboratory, University of Toronto Institute for Aerospace Studies, Canada
}
\date{}

\begin{document}

\maketitle

\begin{abstract}
    \normalsize
Long-distance driving is an important component of planetary surface exploration.
Unforeseen events often require human operators to adjust mobility plans, but this approach does not scale and will be insufficient for future missions.
Interest in self-reliant rovers is increasing, however the research community has not yet given significant attention to autonomous, adaptive decision-making.
    In this paper, we look back at specific planetary mobility operations where human-guided adaptive planning played an important role in mission safety and productivity.
    Inspired by the abilities of human experts, we identify shortcomings of existing autonomous mobility algorithms for robots operating in off-road environments like planetary surfaces.
    We advocate for adaptive decision-making capabilities such as unassisted learning from past experiences and more reliance on stochastic world models. %
    The aim of this work is to highlight promising research avenues to enhance ground planning tools and, ultimately, long-range autonomy algorithms on board planetary rovers.
\end{abstract}

\section{Introduction}

The ability to drive long distances (i.e., on the scale of kilometres) increases the science return of planetary surface missions in many ways.
For instance, it allows rovers to reach regions where landing is prohibitively challenging, such as high elevation martian terrain~\cite{grant_science_2018} or lunar permanently shadowed regions~\cite{owens_development_2021}.
Additionally, this capability enables the collection of close-up surface observations over large areas~\cite{keane_endurance_2022},~\cite[Section 7]{matthies_prospects_2022}.

Long-term surface exploration requires extensive ground-in-the-loop (GITL) operations because current robotic rovers have limited onboard autonomy.
Although human expert oversight is crucial for distilling large amounts of data and ensuring rover safety, it often comes at the cost of productivity.
The operational effectiveness of three Mars Science Laboratory (MSL) science campaigns was studied in~\cite{gaines_productivity_2016}, where 48\% of the sols were classified as ``low productivity,'' meaning that they did not significantly contribute towards the mission objectives.
Preliminary field experiments have empirically shown that greater levels of onboard autonomy would considerably alleviate these productivity issues~\cite{gaines_self-reliant_2020}.
More recently, Sun et al.~\cite{sun_evolution_2024} estimated that nearly 50\% of sols in the first three science campaigns of the Mars 2020 (M2020) mission were not ``campaign-advancing.''
Long-term autonomous planetary surface exploration, which consists of collecting scientific observations over large, unvisited areas with little GITL support, is thus a highly desired and promising capability.

The autonomous capabilities of current rovers are especially beneficial for mobility tasks~\cite{verma_autonomous_2023}.
Yet, even for navigation activities alone, human operator guidance is required on a regular basis. %
The shortcomings in autonomy are often attributed to the underpowered computational hardware on board spacecraft or the bandwidth constraints of deep space communications. %
A more important limitation, however, is that many human operator functions cannot yet be artificially replicated; long-range mobility autonomy in planetary environments is a multifaceted, open problem.
Challenges include scientific data processing to determine global mobility targets, semantic scene understanding, and non-geometric hazard identification, to name a few.
One human operator skill that has received little attention thus far is the ability to characterize traversability uncertainty and adjust navigation plans accordingly. %
Adaptive mobility decision-making involves more than just \textit{reacting} to unforeseen events---it is well-known that mission operators also use their expert judgment to \textit{proactively} assess the safety of future traverses~\cite{biesiadecki_tradeoffs_2007}.

In this paper, we underline the importance of adaptive decision-making in achieving long-range autonomous mobility on planetary surfaces.
In \Cref{sec:flown-missions}, we illustrate how human experts have leveraged adaptive decision-making during previous, flown rover missions.
Then, in \Cref{sec:recommendations}, we identify relevant research work and recommend general capabilities to help bridge the gap between autonomous mobility algorithms and human experts.
We conclude our analysis in~\Cref{sec:conclusion} by briefly mentioning adaptive decision-making mechanisms not explored in this paper.

\section{Adaptive Decision-Making in Flown Rover Missions}
\label{sec:flown-missions}
This section revisits some planetary mobility challenges that were resolved in part through human-guided adaptive behaviours.
We focus on crewed and robotic NASA missions, specifically the Apollo missions, MSL, and M2020.

\subsection{Apollo Missions~\cite{jones_apollo_1995}}

Apollo 15, 16, and 17, referred to as the program's ``J-type'' missions, were the only crewed expeditions involving the lunar roving vehicle (LRV) and took place more than five decades ago.
Ironically, these rover ``systems'' remain the most autonomous ones ever flown as they benefited from the most capable onboard computer for planetary mobility tasks: the human brain.
The adaptive actions of astronauts in response to unforeseen mobility events (on the LRV or on foot), as obvious as some of them might seem, provide a glimpse into the capabilities required for long-distance surface navigation autonomy.

The ground operational teams of the J-type missions had limited information regarding terrain traversability and so contingency driving plans had to be prepared.
For example, the Apollo 15 route planners could not rule out the possibility that the regions to be explored during the first and second extravehicular activities (EVAs) were obstructed with large boulders.
Supplementary terrain traversability assessment procedures were added to the schedule, such as a post-landing ``stand up'' EVA and opportunistic visual inspections during the first EVA.
The Apollo 16 astronauts, on the other hand, had to contend with more rugged terrain and at times, poor lighting conditions.
To make matters worse, mission geologists warned against specific mobility hazards like metres-high scarps along their traverse, but no such features were encountered.
The crew eventually learned what dangers to look out for and adjusted their driving style.
Since finding their way through obstacle fields was a time-consuming task, they would avoid unfamiliar terrain and simply follow outbound rover tracks whenever possible. %

Every crew undoubtedly became more comfortable with driving on the lunar surface as their mission went on.
Lessons learned were also transferred from one mission to the next.
On the hummocky slopes of Hadley Delta, the Apollo 15 crew was surprised by highly unconsolidated soil and one of the astronauts held the LRV in place to prevent any downhill slippage.
Keeping this event in mind, the Apollo 16 crew took the time to park the LRV in a nearby small crater/depression after arriving at a science station on the slopes of Stone Mountain.
Additionally, unlike the previous mission, the Apollo 16 astronauts would not hesitate to lift and carry the LRV to more horizontal grounds before climbing on it if was inconveniently parked. 
The Apollo 17 astronauts completed the program's most ambitious traverses, in part because of the overall confidence in the LRV's capability to handle long drives and steep climbs.
Risk-mitigating walkback and rideback constraints remained one of the points of contention between mission operators and crew members: some astronauts, including the Apollo 17 lunar module pilot Harrison Schmitt, claimed they were too conservative.
Nevertheless, ground teams remained accommodating by constantly adjusting the time allocated for the activities along the traverses in real time.

\subsection{MSL and M2020}

Unlike crewed rover missions, during which complex decisions are made in-situ, robotic missions rely primarily on ground-based operations for safe and adaptive traverse planning.
With the MSL and M2020 missions, global (strategic) traverse route candidates are manually drafted by human experts based on visual assessments of high-resolution orbital data.\footnote{\label{footnote:ops}Based on conversations with MSL and M2020 operations personnel.}
Similarly, human planners evaluate local terrain traversability using rover images and telemetry on a regular basis for local (tactical) planning.
The rovers, on the other hand, are only able to identify geometric hazards and find obstacle-free local paths~\cite{verma_autonomous_2023}.
Thus, human operators must use their expert judgement to set safety limits for a local traverse, including no-go zones, maximum rover slip, time limits, bogie angle thresholds, actuator current limits and more~\cite{rankin_mars_2021}.

At the strategic operational level, notable displays of adaptive route planning on MSL occurred in response to Curiosity's unexpectedly high wheel wear early on in the mission.
Mitigation procedures included simply avoiding regions containing the type of sharp embedded rocks that caused the damage.
Although these dangerous features were too small to be detected on orbital maps, operators related wheel damage with global geomorphic maps to find safe strategic paths~\cite{arvidson_relating_2017}.
Soft terrains were generally preferred over rocky ones to reduce concentrated dynamic loads on the rover's wheels as much as possible, which eventually led to other issues.
On mission sol 709, Curiosity engaged in a traverse across a large ripple field named ``Hidden Valley.''
The slip observed at this location was unexpectedly greater than what was previously experienced on other ripples~\cite{arvidson_mars_2017}.
It took a few sols to return to the entrance of Hidden Valley and it was decided that large ripple fields would be avoided in the future.
Occasionally, when entering a new geologic formation, short exploratory drives would help operators characterize the traversability of the region.\cref{footnote:ops}
An overview of the mobility achievements and challenges during the first seven years of the MSL mission is presented in~\cite{rankin_mars_2021}.
The authors acknowledge the importance of human planner guidance and oversight to mitigate mobility-related risks.

Early into the M2020 mission, it was decided that Perseverance would not commit to a kilometre-long traverse through a rugged formation named Séítah.
Instead, the rover engaged in a traverse that was physically twice as long on more navigable terrain going around Séítah.
Despite the longer distance, it was estimated that the rover's autonomous navigation software would enable a faster traverse to the destination.
The confidence of human operators in this strategy primarily came from older, successful autonomous drives on terrain similar to what the rover would encounter~\cite{rankin_perseverance_2023}.
As with previous missions, long-term drive performance predictions improved as the operational crew accumulated driving experience~\cite{sun_evolution_2024}.

\section{Relevant Work and Open Problems}
\label{sec:recommendations}

This section makes the case that long-range autonomous mobility, in the context of planetary exploration, requires two broad adaptive decision-making mechanisms: unassisted learning from past experiences, and exploiting stochastic rover-environment interaction models.
Inspired by the events enumerated in \Cref{sec:flown-missions}, we identify key capabilities, relevant literature, and challenges that remain unaddressed.

\subsection{Unassisted Learning from Past Experiences}

An obvious trend observed in all rover missions is an increase in familiarity with the environment over time.
Already established as a desired function for autonomy in space~\cite{nesnas_autonomy_2021}, \textit{learning from past experiences} is crucial for planetary surface exploration; unlike other types of spacecraft, rovers constantly operate in partially-known environments filled with hazards.
Mobility plan adjustments are therefore required on a regular basis.
To be effective, an adaptive long-range navigation framework must self-supervise, refining its behaviour based on past traverse performance directly retrieved from telemetry data.
Just as strategic planning experts do in current missions, traverse performance should be associated with the corresponding global environmental context provided by aerial or orbital maps and the general rover state.

The bulk of off-road mobility autonomy research uses slip or traction as a traversability indicator.
In~\cite{endo_deep_2024}, a neural network-based method predicts slip distributions using overhead mapping information like terrain appearance and geometry.
Then, a heuristic function assigns a cost to different parts of the environment, and a safe path is calculated.
At test time, actual slip observations are fed back into the framework to refine predictions over time.
In~\cite{inotsume_terrain_2022}, Gaussian process regression models are fit to previously acquired slip datasets for various base terrain types.
Transfer learning enables traversability inference for new terrain types.
The authors of~\cite{eder_traversability_2023} relate observed traverse costs to different terrain slopes and categories as they appear on elevation and semantic aerial maps.
These costs are interpolated for terrain slope-category pairs not observed in the training data.
The off-road navigation framework in~\cite{cai_evora_2024} leverages deep learning to predict traction distributions for different terrain labels and geometries, in addition to estimating inference reliability.
The approach presented in~\cite{frey_roadrunner_2024} trains a deep traversability estimator directly from rover images and point cloud data collected over kilometres of driving.
Although most field-tested off-road driving frameworks have been demonstrated in (short-range) local planning settings, extensions to (long-range) global mobility are often possible.

An assumption common to most of these approaches is the availability of background knowledge about the environment to assist the learning process. 
For instance, training data collection and test runs in~\cite{cai_evora_2024,frey_roadrunner_2024} were carried out in similar ecoregions.
Additionally, a majority of the referenced work assumes that terrain category maps are available.
Such semantic labels can easily be obtained for off-road mobility on Earth or on Mars, which operators now have decades of experience with~\cite{ono_mars_2018}.
This information is however not available for most bodies in the solar system: Jovian and Saturnian ocean worlds have not been mapped at a high spatial resolution and their surface properties are mostly unknown~\cite{sherwood_program_2018,vaquero_eels_2024}.
Reliance on human-guided labelling would also not be practical for far away destinations like celestial bodies in the Kuiper belt.
Instead, adaptive decision-making frameworks must be able to learn from past experiences \textit{without} external assistance.
Surface mobility performance should be directly related to observable global environmental features like the rover state or source orbital mapping data (such as terrain colour, albedo, elevation, slope, aspect or terrain roughness).
Orbital measurements traditionally reserved for scientific purposes (like hyperspectral imagery, from which thermal inertia can be derived~\cite{cunningham_improving_2019}) should also be considered for adaptive traverse planning and analysis.
Unlike (discrete) human-assigned semantic labels, such quantities are continuous and admit similarity estimation using, for example, covariance functions~\cite{endo_active_2022,inotsume_terrain_2022}.
In turn, quantitative comparisons between global regions of the environment allow more natural traverse predictions in yet unvisited areas.

Learning from past experiences also implies an understanding of what previous mobility measurements, if any, are worth conditioning upcoming decisions upon.
Among the methods mentioned above, those that are aware of training data coverage simply avoid unfamiliar regions of the environment.
Although such a risk-averse posture generally makes sense in safety-critical settings, occasional exploratory drives, as carried out with MSL, can be insightful.
In~\cite{endo_active_2022}, an active learning framework incrementally relates slip observations to terrain slope magnitudes.
In real missions, however, long traverses aim to reach a specific target region; knowledge acquisition is usually not the primary objective.
Exploratory detours should only occur if they might improve future mobility decisions.
This challenge is difficult to tackle with deep learning-based traversability estimation; the aforementioned work in~\cite{cai_evora_2024} only detects out-of-distribution parts of the environment and does not evaluate their utility for future drives.
Once again, surface mobility in exploration settings requires operating beyond the context provided by background traversability knowledge.
As demonstrated with previous missions, rover-terrain interaction models might vary from one region of the environment to the next. Exploratory behaviours can also serve the purpose of updating old, outdated traversability information.

\subsection{Exploiting Stochastic Rover-Terrain Interaction Models}

The sources of uncertainty that characterize partially known environments are generally described as either \textit{epistemic} or \textit{aleatoric}.
Epistemic uncertainty is mitigated through data gathering; the act of learning from past experiences, discussed previously, targets this type of uncertainty.
Aleatoric uncertainty, which represents disturbances caused by unmodelled factors, remains unaffected by observations of the environment.
On planetary surfaces, for instance, different drive outcomes with seemingly identical rover-terrain configurations are caused by aleatoric disturbances.

A stochastic representation of the rover-terrain interaction model enables nuanced environment traversability characterizations and comprehensive adaptive decision-making paradigms. 
The research efforts mentioned in the previous section already express traversability performance probabilistically.
Most of these approaches, however, convert this environment representation into a deterministic one to ease the planning process and represent solutions as paths (a deterministic sequence of actions or successive states).
Nevertheless, some level of adaptability is possible through repeated online calls to the planner in~\cite{endo_deep_2024} and with receding-horizon methods in~\cite{cai_evora_2024,frey_roadrunner_2024}.
For strategic long-range mobility planning, however, policy-based methods are more detailed adaptive planning strategies as they inherently account for future disturbances and corresponding fallback behaviours.
For example, the Canadian Traveller Problem targets epistemic uncertainty and is solved with a policy guiding an agent through an uncertain environment.
In~\cite{guo_robust_2019}, a risk-averse variant of this problem is applied to off-road mobility.
Policy-based methods have also been proposed for strategic rover mission planning problems affected by aleatoric disturbances.
The work in~\cite{lamarre_recovery_2023} tackles a stochastic reachability problem applied to long-range mobility at the lunar south pole with solar-powered rovers.
Furthermore, chance-constrained optimization problems for planetary surface exploration have been solved with policy trees~\cite{santana_risk-aware_2016,lamarre_safe_2024}.

Although stochastic rover-terrain interaction models have the potential to enhance adaptive decision-making schemes for autonomous long-range planetary navigation, many challenges remain.
Most probabilistic models only capture a very small subset of the actual disturbances affecting mobility performance.
For example, the authors of~\cite{lamarre_safe_2024} only account for similar recurring random faults affecting a solar-powered rover at the lunar south pole.
In reality, for such a mission, many more uncertain quantities should be modelled, such as traverse start times, power draw, and effective driving speeds, among others~\cite{shirley_overview_2022}.
Additionally, stochastic reachability analyses and related chance-constrained optimization formulations require clear definitions of rover safety.
The binary interpretation of solar-powered rover safety adopted in~\cite{lamarre_recovery_2023} is overly simplified. 
A more nuanced interpretation, including recovery behaviours in the event of constraint violation, would be needed.
As mentioned in \Cref{sec:flown-missions}, human experts excel at identifying mobility hazards and working towards mission objectives in a risk-averse manner.
Defining and justifying sensible risk-averse postures, such as the Apollo J-type traverse constraints or the strategic traverse plans of martian rover missions, is difficult to accomplish with current algorithms.
Policy-based mobility planning using risk functionals, as demonstrated in~\cite{guo_robust_2019} for off-road mobility, is an emerging and promising field of study on this end~\cite{wang_risk-averse_2022}.

\section{Conclusion}
\label{sec:conclusion}

In this work, we highlighted a clear disparity between human-guided and autonomous adaptive decision-making for long-range planetary surface mobility.
We identified two broad autonomy algorithm improvements that would help to bridge this gap: unassisted learning from past experiences and taking more advantage of stochastic rover-terrain interaction models.
In addition to increasing mission productivity and enabling the exploration of distant extra-terrestrial worlds, enhanced onboard autonomy would lead to better human operator-rover symbiosis~\cite{nesnas_autonomy_2021}.

This paper has only scratched the surface of the challenges associated with long-range adaptive surface mobility.
For example, \textit{explainable adaptation} is important.
Apollo astronauts would constantly update mission control about unforeseen events and the rationale behind their decision-making, which in turn helped ground operators provide useful feedback. 
By the same token, explainable adaptive mobility algorithms would provide greater context to robotic mission operators.
At the moment, MSL and M2020 ground teams only develop their navigation planning expertise through the telemetry that they themselves explicitly queue for downlink.

Another under-explored challenge related to long-range surface autonomy is the locomotion hardware, which must \textit{physically enable adaptation}.
The autonomous navigation capabilities of current rovers are, in part, limited by the potential consequences of a bad mobility decision. 
With MSL and M2020, traversing fields of megaripples or terrain covered in large boulders could lead to rover entrapment and, effectively, the end of their intended mission.
The lack of backup mobility modes discourages driving on unfamiliar terrain.
Rover platforms equipped with more resilient hardware designs, such as wheeled-legged hybrid locomotion systems, are favourable to information gathering and more efficient traverses in the long term.

\sloppy
\printbibliography

@article{arvidson_mars_2017,
  author = {Arvidson, Raymond E. and Iagnemma, Karl D. and Maimone, Mark and Fraeman, Abigail A. and Zhou, Feng and Heverly, Matthew C. and Bellutta, Paolo and Rubin, David and Stein, Nathan T. and Grotzinger, John P. and Vasavada, Ashwin R.},
  copyright = {© 2016 Wiley Periodicals, Inc.},
  doi = {https://doi.org/10.1002/rob.21647},
  issn = {1556-4967},
  journal = {J. Field Robot.},
  language = {en},
  number = {3},
  pages = {495--518},
  title = {Mars {Science} {Laboratory} {Curiosity} {Rover} {Megaripple} {Crossings} up to {Sol} 710 in {Gale} {Crater}},
  urldate = {2021-01-25},
  volume = {34},
  year = {2017}
}

@article{arvidson_relating_2017,
  author = {Arvidson, R. E. and DeGrosse, P. and Grotzinger, J. P. and Heverly, M. C. and Shechet, J. and Moreland, S. J. and Newby, M. A. and Stein, N. and Steffy, A. C. and Zhou, F. and Zastrow, A. M. and Vasavada, A. R. and Fraeman, A. A. and Stilly, E. K.},
  doi = {10.1016/j.jterra.2017.03.001},
  issn = {0022-4898},
  journal = {J. Terramechanics},
  language = {en},
  month = {Oct.},
  pages = {73--93},
  series = {Manned/{Unmanned} {Ground} {Vehicles}: {Off}-{Road} {Dynamics} and {Mobility}},
  title = {Relating geologic units and mobility system kinematics contributing to {Curiosity} wheel damage at {Gale} {Crater}, {Mars}},
  urldate = {2021-01-22},
  volume = {73},
  year = {2017}
}

@article{biesiadecki_tradeoffs_2007,
  author = {Biesiadecki, Jeffrey J. and Leger, P. Chris and Maimone, Mark W.},
  doi = {10.1177/0278364907073777},
  issn = {0278-3649},
  journal = {Int. J. Robot. Res.},
  language = {en},
  month = {Jan.},
  number = {1},
  pages = {91--104},
  title = {Tradeoffs {Between} {Directed} and {Autonomous} {Driving} on the {Mars} {Exploration} {Rovers}},
  urldate = {2023-10-17},
  volume = {26},
  year = {2007}
}

@article{cai_evora_2024,
  author = {Cai, Xiaoyi and Ancha, Siddharth and Sharma, Lakshay and Osteen, Philip R. and Bucher, Bernadette and Phillips, Stephen and Wang, Jiuguang and Everett, Michael and Roy, Nicholas and How, Jonathan P.},
  doi = {10.1109/TRO.2024.3431828},
  issn = {1941-0468},
  journal = {IEEE Trans. Robot.},
  pages = {3756--3777},
  shorttitle = {{EVORA}},
  title = {{EVORA}: {Deep} {Evidential} {Traversability} {Learning} for {Risk}-{Aware} {Off}-{Road} {Autonomy}},
  urldate = {2024-09-09},
  volume = {40},
  year = {2024}
}

@article{cunningham_improving_2019,
  author = {Cunningham, Christopher and Nesnas, Issa A. and Whittaker, William L.},
  doi = {10.1007/s10514-018-9796-4},
  issn = {1573-7527},
  journal = {Autonomous Robots},
  language = {en},
  month = {Feb.},
  number = {2},
  pages = {503--521},
  title = {Improving slip prediction on {Mars} using thermal inertia measurements},
  urldate = {2024-09-10},
  volume = {43},
  year = {2019}
}

@article{eder_traversability_2023,
  author = {Eder, Matthias and Prinz, Raphael and Schöggl, Florian and Steinbauer-Wagner, Gerald},
  doi = {10.1016/j.robot.2023.104494},
  issn = {0921-8890},
  journal = {Robot. and Autonomous Syst.},
  month = {Oct.},
  pages = {104494},
  title = {Traversability analysis for off-road environments using locomotion experiments and earth observation data},
  urldate = {2024-09-09},
  volume = {168},
  year = {2023}
}

@article{endo_active_2022,
  author = {Endo, Masafumi and Ishigami, Genya},
  doi = {10.1109/LRA.2022.3207554},
  issn = {2377-3766},
  journal = {IEEE Robot. and Autom. Letters},
  month = {Oct.},
  number = {4},
  pages = {11855--11862},
  title = {Active {Traversability} {Learning} via {Risk}-{Aware} {Information} {Gathering} for {Planetary} {Exploration} {Rovers}},
  volume = {7},
  year = {2022}
}

@article{endo_deep_2024,
    title={Deep Probabilistic Traversability with Test-time Adaptation for Uncertainty-aware Planetary Rover Navigation},
    author={Endo, Masafumi and Taniai, Tatsunori and Ishigami, Genya},
    journal={arXiv preprint arXiv:2409.00641},
    year={2024}
}

@article{frey_roadrunner_2024,
  title={RoadRunner--Learning Traversability Estimation for Autonomous Off-road Driving},
  author={Frey, Jonas and Khattak, Shehryar and Patel, Manthan and Atha, Deegan and Nubert, Julian and Padgett, Curtis and Hutter, Marco and Spieler, Patrick},
  journal={arXiv preprint arXiv:2402.19341},
  year={2024}
}

@inproceedings{gaines_productivity_2016,
  author = {Gaines, Daniel and Anderson, Robert and Doran, Gary and Huffman, William and Justice, Heather and Mackey, Ryan and Rabideau, Gregg and Vasavada, Ashwin and Verma, Vandana and Estlin, Tara},
  booktitle = {Proc. 4th {Workshop} on {Planning} and {Robot.}},
  pages = {115--125},
  publisher = {London, UK},
  title = {Productivity challenges for mars rover operations},
  urldate = {2023-10-31},
  year = {2016}
}

@article{gaines_self-reliant_2020,
  author = {Gaines, Daniel and Doran, Gary and Paton, Michael and Rothrock, Brandon and Russino, Joseph and Mackey, Ryan and Anderson, Robert and Francis, Raymond and Joswig, Chet and Justice, Heather and Kolcio, Ksenia and Rabideau, Gregg and Schaffer, Steve and Sawoniewicz, Jacek and Vasavada, Ashwin and Wong, Vincent and Yu, Kathryn and Agha‐mohammadi, Ali-akbar},
  doi = {10.1002/rob.21979},
  issn = {1556-4967},
  journal = {J. Field Robot.},
  language = {en},
  month = {Oct.},
  number = {7},
  pages = {1171--1196},
  title = {Self-reliant rovers for increased mission productivity},
  urldate = {2020-08-22},
  volume = {37},
  year = {2020}
}

@article{grant_science_2018,
  author = {Grant, John A. and Golombek, Matthew P. and Wilson, Sharon A. and Farley, Kenneth A. and Williford, Ken H. and Chen, Al},
  doi = {10.1016/j.pss.2018.07.001},
  issn = {0032-0633},
  journal = {Planet. and Space Sci.},
  month = {Dec.},
  pages = {106--126},
  title = {The science process for selecting the landing site for the 2020 {Mars} rover},
  urldate = {2023-10-24},
  volume = {164},
  year = {2018}
}

@inproceedings{guo_robust_2019,
  author = {Guo, H. and Barfoot, T. D.},
  booktitle = {2019 {Int.} {Conf.} on {Robot.} and {Autom.}},
  doi = {10.1109/ICRA.2019.8794252},
  month = {May},
  pages = {5523--5529},
  title = {The {Robust} {Canadian} {Traveler} {Problem} {Applied} to {Robot} {Routing}},
  year = {2019}
}

@article{inotsume_terrain_2022,
  author = {Inotsume, Hiroaki and Kubota, Takashi},
  doi = {10.1186/s40648-021-00215-3},
  issn = {2197-4225},
  journal = {ROBOMECH J.},
  month = {Feb.},
  number = {1},
  pages = {6},
  title = {Terrain traversability prediction for off-road vehicles based on multi-source transfer learning},
  urldate = {2022-03-02},
  volume = {9},
  year = {2022}
}

@misc{jones_apollo_1995,
  author = {Jones, Eric M.},
  title = {Apollo {Lunar} {Surface} {Journal}},
  note = {URL: www.nasa.gov/history/alsj (visited on 08/01/2024)},
  urldate = {2024-08-01},
  year = {1995}
}

@inproceedings{keane_endurance_2022,
  author = {Keane, J. T. and Tikoo, S. M. and Elliott, J.},
  booktitle = {Lunar {Exploration} {Analysis} {Group} meeting},
  shorttitle = {Endurance},
  title = {Endurance: lunar south pole-aitken basin traverse and sample return rover},
  year = {2022}
}

@article{lamarre_recovery_2023,
  author = {Lamarre, Olivier and Malhotra, Shantanu and Kelly, Jonathan},
  doi = {10.1016/j.actaastro.2023.09.028},
  issn = {0094-5765},
  journal = {Acta Astronaut.},
  month = {Dec.},
  pages = {708--724},
  title = {Recovery policies for safe exploration of lunar permanently shadowed regions by a solar-powered rover},
  urldate = {2023-11-18},
  volume = {213},
  year = {2023}
}

@inproceedings{lamarre_safe_2024,
  author = {Lamarre, Olivier and Malhotra, Shantanu and Kelly, Jonathan},
  booktitle = {2024 {IEEE} {Aerospace} {Conf.}},
  doi = {10.1109/AERO58975.2024.10521136},
  month = {Mar.},
  pages = {1--14},
  title = {Safe {Mission}-{Level} {Path} {Planning} for {Exploration} of {Lunar} {Shadowed} {Regions} by a {Solar}-{Powered} {Rover}},
  year = {2024}
}

@inproceedings{matthies_prospects_2022,
  author = {Matthies, Larry and Kennett, Andrew and Kerber, Laura and Fraeman, Abigail and Anderson, Robert C.},
  booktitle = {2022 {IEEE} {Aerospace} {Conf.}},
  doi = {10.1109/AERO53065.2022.9843681},
  month = {Mar.},
  pages = {1--11},
  title = {Prospects for {Very} {Long}-{Range} {Mars} {Rover} {Missions}},
  urldate = {2024-04-06},
  year = {2022}
}

@article{nesnas_autonomy_2021,
  author = {Nesnas, Issa A.D. and Fesq, Lorraine M. and Volpe, Richard A.},
  doi = {10.1007/s43154-021-00057-2},
  issn = {2662-4087},
  journal = {Current Robot. Reports},
  language = {en},
  month = {June},
  shorttitle = {Autonomy for {Space} {Robots}},
  title = {Autonomy for {Space} {Robots}: {Past}, {Present}, and {Future}},
  urldate = {2021-06-29},
  year = {2021}
}

@inproceedings{ono_mars_2018,
  author = {Ono, Masahiro and Heverly, Matthew and Rothrock, Brandon and Almeida, Eduardo and Calef, Fred and Soliman, Tariq and Williams, Nathan and Gengl, Hallie and Ishimatsu, Takuto and Nicholas, Austin},
  booktitle = {2018 {AIAA} {SPACE} and {Astronautics} {Forum} and {Expo.}},
  pages = {5419},
  shorttitle = {Mars 2020 site-specific mission performance analysis},
  title = {Mars 2020 site-specific mission performance analysis: {Part} 2. {Surface} traversability},
  year = {2018}
}

@incollection{owens_development_2021,
  author = {Owens, Chris and Macdonald, Kori and Hardy, Jeremy and Lindsay, René and Redfield, Morgan and Bloom, Michael and Bailey, Erik and Cheng, Yang and Clouse, Dan and Villalpando, Carlos Y. and Hambardzumyan, Ashot and Johnson, Andrew E. and Horchler, Andrew D.},
  booktitle = {{AIAA} {Scitech} 2021 {Forum}},
  doi = {10.2514/6.2021-0376},
  month = {Jan.},
  publisher = {American Institute of Aeronautics and Astronautics},
  title = {Development of a {Signature}-based {Terrain} {Relative} {Navigation} {System} for {Precision} {Landing}},
  urldate = {2023-10-25},
  year = {2021}
}

@article{rankin_mars_2021,
  author = {Rankin, Arturo and Maimone, Mark and Biesiadecki, Jeffrey and Patel, Nikunj and Levine, Dan and Toupet, Olivier},
  copyright = {© 2021 Wiley Periodicals LLC},
  doi = {https://doi.org/10.1002/rob.22011},
  issn = {1556-4967},
  journal = {J. Field Robot.},
  language = {en},
  month = {Aug.},
  number = {5},
  pages = {759--800},
  title = {Mars curiosity rover mobility trends during the first 7 years},
  urldate = {2021-01-18},
  volume = {38},
  year = {2021}
}

@inproceedings{rankin_perseverance_2023,
  author = {Rankin, Arturo and Del Sesto, Tyler and Hwang, Pauline and Justice, Heather and Maimone, Mark and Verma, Vandi and Graser, Evan},
  booktitle = {2023 {IEEE} {Aerospace} {Conf.}},
  doi = {10.1109/AERO55745.2023.10115835},
  month = {Mar.},
  pages = {1--16},
  title = {Perseverance {Rapid} {Traverse} {Campaign}},
  urldate = {2023-10-18},
  year = {2023}
}

@inproceedings{santana_risk-aware_2016,
  author = {Santana, Pedro and Vaquero, Tiago and McGhan, Catharine L. and Toledo, Claudio and Timmons, Eric and Williams, Brian and Murray, Richard},
  booktitle = {{AIAA} {SPACE} 2016},
  doi = {10.2514/6.2016-5537},
  pages = {5537},
  shorttitle = {Risk-aware planning in hybrid domains},
  title = {Risk-aware planning in hybrid domains: {An} application to autonomous planetary rovers},
  year = {2016}
}

@article{sherwood_program_2018,
  author = {Sherwood, B. and Lunine, J. and Sotin, C. and Cwik, T. and Naderi, F.},
  doi = {10.1016/j.actaastro.2017.11.047},
  issn = {0094-5765},
  journal = {Acta Astronaut.},
  month = {Feb.},
  pages = {285--296},
  title = {Program options to explore ocean worlds},
  urldate = {2024-09-06},
  volume = {143},
  year = {2018}
}

@techreport{shirley_overview_2022,
  author = {Shirley, Mark and Balaban, Edward},
  institution = {National Aeronautics and Space Administration},
  number = {20220008301},
  title = {An {Overview} of {Mission} {Planning} for the {VIPER} {Rover}},
  type = {Presentation},
  year = {2022}
}

@inproceedings{sun_evolution_2024,
  author = {Sun, Vivian Z. and Sholes, Steven and Stack Morgan, Katie and Farley, Ken and Del Sesto, Tyler and Kronyak, Rachel and Pyrzak, Guy and Welch, Richard and Lange, Robert},
  booktitle = {2024 {IEEE} {Aerospace} {Conf.}},
  doi = {10.1109/AERO58975.2024.10521069},
  month = {Mar.},
  pages = {1--16},
  title = {Evolution of the {Mars} 2020 {Perseverance} {Rover}’s {Strategic} {Planning} {Process}},
  year = {2024}
}

@article{vaquero_eels_2024,
  author = {Vaquero, T. S. and Daddi, G. and Thakker, R. and Paton, M. and Jasour, A. and Strub, M. P. and Swan, R. M. and Royce, R. and Gildner, M. and Tosi, P. and Veismann, M. and Gavrilov, P. and Marteau, E. and Bowkett, J. and de Mola Lemus, D. Loret and Nakka, Y. and Hockman, B. and Orekhov, A. and Hasseler, T. D. and Leake, C. and Nuernberger, B. and Proença, P. and Reid, W. and Talbot, W. and Georgiev, N. and Pailevanian, T. and Archanian, A. and Ambrose, E. and Jasper, J. and Etheredge, R. and Roman, C. and Levine, D. and Otsu, K. and Yearicks, S. and Melikyan, H. and Rieber, R. R. and Carpenter, K. and Nash, J. and Jain, A. and Shiraishi, L. and Robinson, M. and Travers, M. and Choset, H. and Burdick, J. and Gardner, A. and Cable, M. and Ingham, M. and Ono, M.},
  doi = {10.1126/scirobotics.adh8332},
  journal = {Sci. Robot.},
  month = {Mar.},
  number = {88},
  pages = {eadh8332},
  shorttitle = {{EELS}},
  title = {{EELS}: {Autonomous} snake-like robot with task and motion planning capabilities for ice world exploration},
  urldate = {2024-03-16},
  volume = {9},
  year = {2024}
}

@article{verma_autonomous_2023,
  author = {Verma, Vandi and Maimone, Mark W. and Gaines, Daniel M. and Francis, Raymond and Estlin, Tara A. and Kuhn, Stephen R. and Rabideau, Gregg R. and Chien, Steve A. and McHenry, Michael M. and Graser, Evan J. and Rankin, Arturo L. and Thiel, Ellen R.},
  doi = {10.1126/scirobotics.adi3099},
  journal = {Sci. Robot.},
  month = {July},
  number = {80},
  pages = {eadi3099},
  title = {Autonomous robotics is driving {Perseverance} rover’s progress on {Mars}},
  urldate = {2023-08-17},
  volume = {8},
  year = {2023}
}

@article{wang_risk-averse_2022,
  author = {Wang, Yuheng and Chapman, Margaret P.},
  doi = {10.1016/j.artint.2022.103743},
  issn = {0004-3702},
  journal = {Artif. Intell.},
  language = {en},
  month = {Oct.},
  pages = {103743},
  shorttitle = {Risk-averse autonomous systems},
  title = {Risk-averse autonomous systems: {A} brief history and recent developments from the perspective of optimal control},
  urldate = {2022-11-28},
  volume = {311},
  year = {2022}
}

\end{document}